# A Machine Learning Approach for Driver Identification Based on CAN-BUS Sensor Data

Md. Abbas Ali Khan[1], Mohammad Hanif Ali[2], A K M Fazlul Haque[1], Md. Tarek Habib[1]
[1]Daffodil International University Bangladesh, [2]Jahangirnagar University Bangladesh
abbas.cse@diu.edu.bd

**Abstract:**
Driver identification is a momentous field of modern decorated vehicles in the perspective of the controller area network (CAN-BUS). Many conventional systems are used to identify the driver. One step ahead, most of the researchers use sensor data of CAN-BUS but there are some difficulties because of the variation of protocol of different models of vehicle. Our aim is to identify the driver through supervised learning algorithms based on driving behavior analysis. To identify the driver, a driver verification technique is proposed that evaluate driving pattern using the measurement of CAN sensor data. In this paper on-board diagnostic (OBD-II) is used to capture the data from CAN-BUS sensor and the sensors are listed under SAE J1979 statement. According to the service of OBD-II drive identification is possible. However, we have gained two types of accuracy on a full data set with 10 drivers and a partial data set with two drivers. The accuracy is good with less number of drivers compared to higher number of the driver. We have achieved statistically significant results in terms of accuracy contrast to the baseline algorithm.

**Keywords:** Driver Identification, Pattern Analysis, CAN-BUS, OBD-II, Machine Learning.

## 1. Introduction:

Every driver has their own driving style, therefore the driver can be classified according to exploration through the driving pattern analysis. It is to be considered as a fingerprint of the driver's manner like acceleration, speed, braking habits that vary from driver to driver. Driver fingerprinting could lead to important privacy compromises [1].

Today we cannot consider just a vehicle as a modern car, as it is a fully decorated smart device with various functionality like multimedia, security system and different sensors [2]. At most three sensors named fuel level, coolant temperature, and oil pressure were furnished last century until the 70th year. The sensors were very simple because the driver was informed regarding the features of the engine and amount of fuel through the magnetoelectric and light display devices [2]. Nowadays, cars are equipped with many microcomputers. Information technology is developing rapidly and cars are connected with the internet. Using the state-of-the art technology in real time all the microcomputers are communicated to each other through CAN-BUS (Controller Area Network) [3]. As a result the drivers feel secure and joyful of their trips and all other equipment are functioning properly.

To make a car more efficient a good number of technologies are used in the modern engine. To improve the engine performance direct injection technology introduced in the modern car [4]. According to a survey, the researcher predicted that the number of sales of connected cars will reach 76.3 million in the next 2023 [5]. In the near future technology based connected cars will make a digital platform where multitude of sensors will take place like radar, LIDAR, cameras, ultrasonic sensors, and vehicle motion sensors [5]. Through the state-of-the-art technology modern engines use less fuel and besides get more power [6]. Most of the cars have partnered with other components which are highly technology based, such as traffic lights, garage doors and services [7]. Cars on the dashboard have green lights that indicate the drivers' efficient driving. It's improving driving style and fuel consumption. Not only on the driving style there is a discount policy on insurance services but also real time monitoring, maintenance, path finding, driving style development and also consumption of fuel [8].

There are many bugs and vulnerabilities likely in the software operating cars. Because of internet connection exploit vulnerabilities are feasible for the attackers. For collecting data a car cooperates with a third party server and the car becomes more vulnerable to theft. Till now many vulnerabilities are discovered and it's a continuous process. The more technology the car is based on, the more intelligent the thieves are. In the modern era, various modern techniques are used to steal a car key by the attacker. Vulnerabilities of connected cars will increase the auto-theft that is one of the threats [9]. Top-of-the range vehicles are targeted by thieves who simply drive off after bypassing security devices by hacking on-board computers [10].Even the keyless cars, once connected the thieves can access the vehicle's electronic information, allowing them to drive it away. One technique involves breaking into the vehicle and plugging a laptop into the hidden diagnostic socket [10]. Penny [11] introduced man-in-the middle attack or relay attack, to do this radio signals are passed between two devices. Pekaric I at al. (2021) [12] described another attack such as GPS spoofing and message injection attacks. BMW Connected Drive [13] seamlessly integrates mobile devices, smart home technology, and vehicle's intelligent interfaces into a complete driver's environment. Even Though in 2021 they introduced a remote door unlock system through a signal to the driver's door to unlock [13].The threats being discovered will be realized and the security of connected cars will become more important as more cars are connected to the internet [8].

Previous researchers introduced biometric authentication as one of the significant tools based on the physical characteristics of the driver like fingerprint, face or voice detection, eye shell scanning and also behavioral characteristics [2]. Recognizing/analyzing the driver's driving pattern is a salient feature to develop the security of a car. Data-mining techniques are widely used by earlier researchers to detect such a novel attack. Because each driver has their own driving style, data-mining is also a prominent method to detect car theft (due to unexpected driving style). As we say that the basis of telemetric data the features of the driver's driving pattern are reflected.

CAN-BUS is likely a nervous system used to allow configuration, data logging and communication among electronic control units (ECU) e.g. ECU is like a part of the body and interconnected through CAN, by which information sensed by one part can be shared with another [14]. Up to 70 ECUs have a modern car e.g. the engine control unit, airbags, audio system, acceleration, fuel unit etc. [15].

Normally, multi-sensors data is made up of in vehicle's CAN data. The in vehicle CAN data such as steering wheel, vehicle speed, engine speed, amount of fuel, etc. Several researchers previously proposed a driver identification method based on in vehicle CAN-BUS data. But direct connectivity is difficult to get data, so on-board diagnostics (ODB-II) is used. (OBD-II, ISO 15765) is a self-diagnostic and reporting capability that e.g. mechanics use to identify car issues, OBD-II specifies diagnostic trouble codes (DTCs) and real-time data (e.g. speed, revolution per minute RPM), which can be recorded via OBD-II loggers from CAN-BUS. Though such data is difficult to get, every moment data is passing, we need the PID (parameter identifier) number of each specific feature to correctly extract. It is non-public and it is made up on the basis of the company. Many authors described the problem of CAN-BUS data for identifying the driver [9], [16], [17].

In this paper, our aim is to identify the driver behavior through telemetric data using machine learning algorithms. We analyze the data in terms of training, testing and validation to get model accuracy that helps us to driver identification.

## 2. Literature Review

The author [18] uses telemetric data to investigate the driver identification and identification accuracy decreases 15% compared to the method. They use the role of non-public parameters in identifying the driver. A previous work had been done by using car driving simulated [18] data. Investigated the driver's behavior when he follows another car. The features mentioned below are used to observe such as accelerator pedal, car speed, brake pedal and distance to the next car. Gaussian Mixture Model (GMM) is used to achieve 81% accuracy with 12 drivers and 73% of 30 drivers [18]. Another author [19] analyzes overtaking style for each driver, uses accelerator, and steering data and the accuracy is 85% about 20 drivers through the Hidden Markov Model (HMM). Some other authors used smartphones to capture driver data. Sensors in the smartphone are- GPS, accelerometer, magnetometer, and gyroscope [20-22]. This data is used for driver profiling and other tasks. An Author used inertial sensor and algorithm was SVM, *k*-means methods and got 60% accuracy between two drivers.

In another research, the authors [8, 16], [23-24] acquire data from in vehicle CAN-BUS via OBD-II. The author [23] uses in vehicle CAN-BUS sensor data and the accuracy was 99% among 15 drivers. He uses SVM, Random Forest, Naive Bayes, and *k*-NN methods. Another researcher [9] got 99% accuracy from 51 features of 10 drivers and used Decision Tree, *k*-NN, Random Forest and Multilayer Perceptron (MPL). Choi et al. [24] find out the driving detection and driver recognition using both GMM and HMM methods, they are used for analyzing in vehicle CAN-BUS data. Kedar-Dongakar et al. [25] recognized the driver classification based on energy optimization of a vehicle. Based on driving style three types of drivers are classified as aggressive, moderate and conservative. The author considers the following features for his research work such as vehicle speed, acceleration, torque, acceleration pedal, and steering wheel angle and brake pedal pressure.

Several researches have been going on neural network and deep learning algorithms for a few years back and draw a good impact on driver behavior identification works. Xun et al. [26] introduced Convolutional Neural Network (CNN) and got 99% accuracy for 10 drivers. For driver identification another paper uses noise free data and as algorithm- LSTM-Recurrent Neural Network is used, where it has high accuracy with salient advantages [27].

## 3. Architecture of the Intended System

An architecture of the intended system is proposed to identify the authorized driver as shown in Figure 1. The modern vehicles are connected to the internet through IEEE 802 standard, transfer the driver data. The analysis module analyze the data. If the driving pattern is not match with the accredited driver then the driver identification cell detects and send message to the owner of the vehicle.

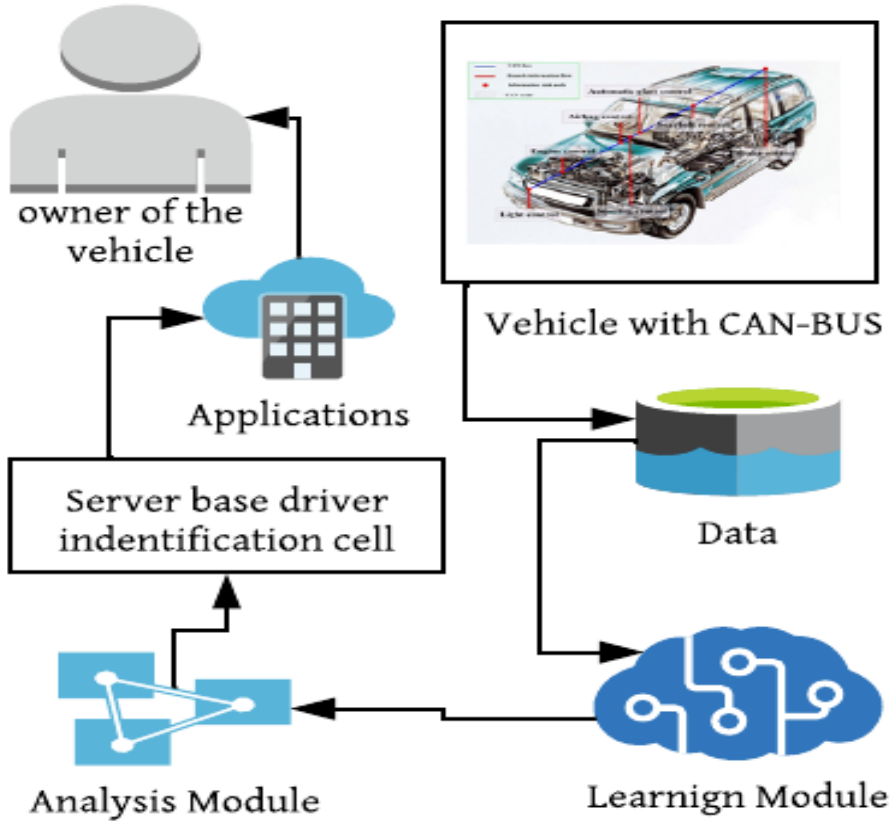

Figure 1 Architecture of the Intended system for driver identification

## 4. Methods

Steps of the proposed system are shown in Figure 2. Until before the classifiers dataset preparation, data preprocessing, feature selection, normalization and other activities are occurred. Ensemble and Baseline classifiers are used for comparative analysis with the state-of-the-art classifier to find the best model.

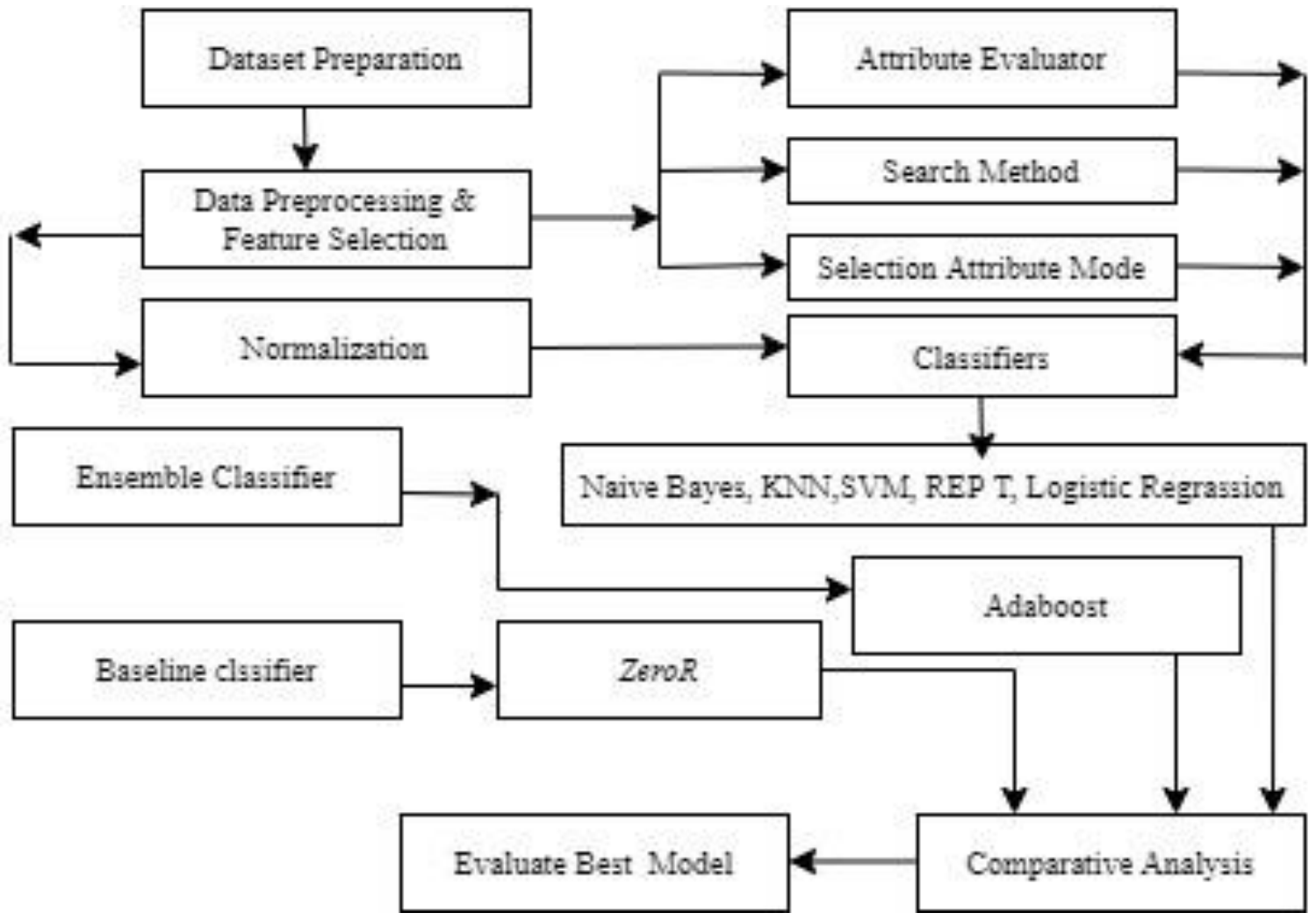

Figure 2 steps of our proposed system

## 4.1 Dataset Preparation

In this connection we need data of the trips for driver identification. Our model is considered an Ocslab driving dataset [28]. This date is used for driver classification and personalization based on pattern analysis. KIA motors corporation vehicles in South Korea were performed to collect the data and the experiment has been done since July 28, 2015. Total 10 drivers labeled “A” to “J” are included in the trips and cover 23 km length, completing two round trips from 8.00 PM to 11.00 PM. Three types (such as city road, freeway, and parking lot) of road are there with their own characteristics. There are a total 94,401 records with 51 dimensions (51 features) and Table 1 depicts the Ocslab dataset.

In real driving condition each driver drove their own style, in-vehicle CAN-BUS data were collected with OBD-II and CarbbigsP (OBD-II scanner). Not all data is possible to get because there are some limitations of OBD-II identifiers and sensors such as it cannot provide body control status or airbag status even wheel angle rotation status. OBD-II has a limited set of identifier [29] provided by the manufacturer, Table 3 shows some list of PIDs (Parameter IDs) of service/mode (Hex) 01. There are 10 diagnostic services described in the latest OBD-II standard SAE J1979 [29]. Few numbers of services are shown in Table 2.

Table 1 Driving dataset of Ocslab

| Type | Features |
|---|---|
| Engine | Engine torque<br>Engine coolant temperature<br>Maximum indicated engine torque<br>Activation of Air Compressor<br>….<br>Friction torque |
| Fuel | Long term fuel trim Bank1<br>Intake air pressure<br>Accelerator pedal value<br>………..<br>Fuel consumption |
| Transmission | Transmission oil temperature<br>Wheel velocity, front,left-hand<br>Wheel velocity, front,right-hand<br>Wheel velocity, rear,left-hand<br>……..<br>Torque converter Speed |

Driver’s driving statistical features are exposed. Figure 3 shows the time series pattern of the in-vehicle's CAN data of in the real time driving situation of driver A and D, where data fluctuation is visible in RPM.

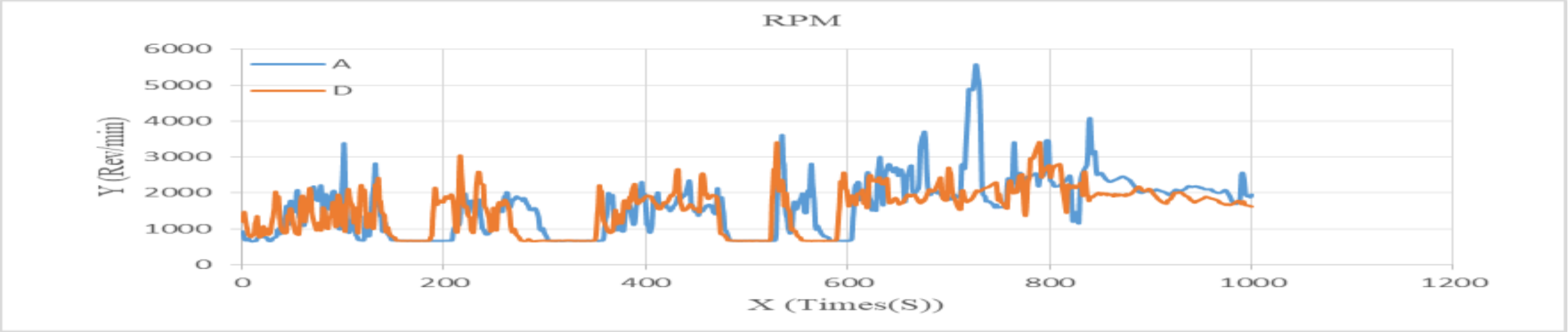

Figure 3. The RPM (Revolutions per Minute) of driver A and D

Table 2 list of some OBD-II services/ mode (hex)

| Service/Mode(Hex) | Description |
|---|---|
| 01 | Show current data |
| 02 | Show freeze frame data |
| …….. | …….. |
| 09 | Request vehicle information |
| 0A | Permanent Diagnostic Trouble Codes |

Table 3 list some OBD-II parameter

| Service /Mode (Hex) | PID(hex) | Data byte returned | Description | Min Value | Max Value | Units |
|---|---|---|---|---|---|---|
| 01 | 03 | 2 | Fuel system status | - | - | - |
| | 04 | 1 | Calculated engine load | 0 | 100 | % |
| | 0C | 2 | Engine speed | 0 | 16,383.75 | rpm |
| | OD | 1 | Vehicle speed | 0 | 255 | km/h |
| | -- | --- | ---- | -- | -- | -- |
| | 68 | 3 | Intake air temperature sensors | -40 | 215 | $^0$C |

## 4.2 Data Prepossessing

There are 51 features used in our work in the dataset. Transform the collected data to our classification model for analysis we follow- feature selection, data normalization and data processing through sliding window technique. The sample data are viewed in equations (1).

$$X = \begin{bmatrix} X_1^1 & X_2^1 \ldots & X_d^1 \\ X_1^2 & X_2^2 \ldots & X_d^2 \\ \vdots & \vdots & \vdots \\ X_1^N & X_2^N \ldots & X_d^N \end{bmatrix} \quad (1)$$

Where *d* columns correspond to *d* variable and *N* rows correspond to *N* instances.

### 4.2.1 Feature Selection

We discard the following kind of features from the dataset for ameliorative achievement and accuracy of the model. We have considered CoorelationAttributeEval as attribute evaluator, Ranker is used for search method and also select cross-validation 10 and seed 1 while selecting the attribute mode.

- Homogeneous feature = $A_h$
- Irrelevant feature = $B_i$
- Superfluous feature and = $C_s$
- Mostly Correlated feature = $D_c$

*Engine_torque* and *correction_of_engine_torque* features are identical as well as *engine_coolant_temperature* is a redundant feature. Hence, selection of features referred to in [2] was performed, from the original dataset of 51 selecting 15 features. Table 4 shows the selected feature with statistical significance of mean and standard deviation equation in (2) and (3) respectively.

$$\bar{x} = \frac{\sum_{i=1}^{N} x_i}{N} \tag{2}$$

$$\sigma = \sqrt{\frac{\sum (x_i - \mu)^2}{N}} \tag{3}$$

Table 4 Selected 15 feature with mean and standard deviation

| Feature | Vehicle Data Type | Mean | Standard deviation | Previous work | Classifiers |
|---|---|---|---|---|---|
| Long term fuel trim bank1 | Fuel | 2.843 | 1.363 | [9] | DT, KNN, RF, MLP |
| Intake air pressure | | 36.85 | 27.95 | | |
| Accelerator pedal value | | 3.719 | 8.506 | [18],[25],[30],[31],[32] | GMM,SMG,MM,GMM, MLP,SM,FNN, |
| Fuel consumption | | 757 | 761.13 | | |
| Maximum indicated engine torque | Engine | 67.5 | 9.5 | | |
| Engine torque’ | | 23.75 | 14.73 | [23] | SVM,RF,NB,KNN |
| Calculated load value | | 41.30 | 18.38 | | |
| Friction torque | | 13.7 | 2.27 | | |
| Activation of air compressor | | 0.89 | 0.31 | | |
| Engine coolant temperature | Transmission | 84.24 | 6.12 | | |
| Transmission oil temperature | | 80.21 | 10.5 | [9] | DT, KNN, RF, MLP |
| Wheel velocity front left-hand | | 30.11 | 26.48 | | |
| Wheel velocity front right-hand | | 29.36 | 26.22 | | |
| Wheel velocity rear left-hand | | 29.20 | 26.10 | | |
| Torque converter speed’ | | 1259.15 | 766.51 | | |

### 4.2.2 Data Normalization

As we see, different scales of data exist in the dataset. So we are to normalize the data according to the min-max approach. Normalization is essential for some machine learning algorithms like *k*NN (*k*-Nearest Neighbor) and SVM [29]. The normalization formulas for integrating data scales are shown in equation 4.

$$X_{norm} = \frac{X - X_{min}}{X_{max} - X_{min}} \tag{4}$$

Here, min means minimum value of a feature and max refers to the maximum value respectively

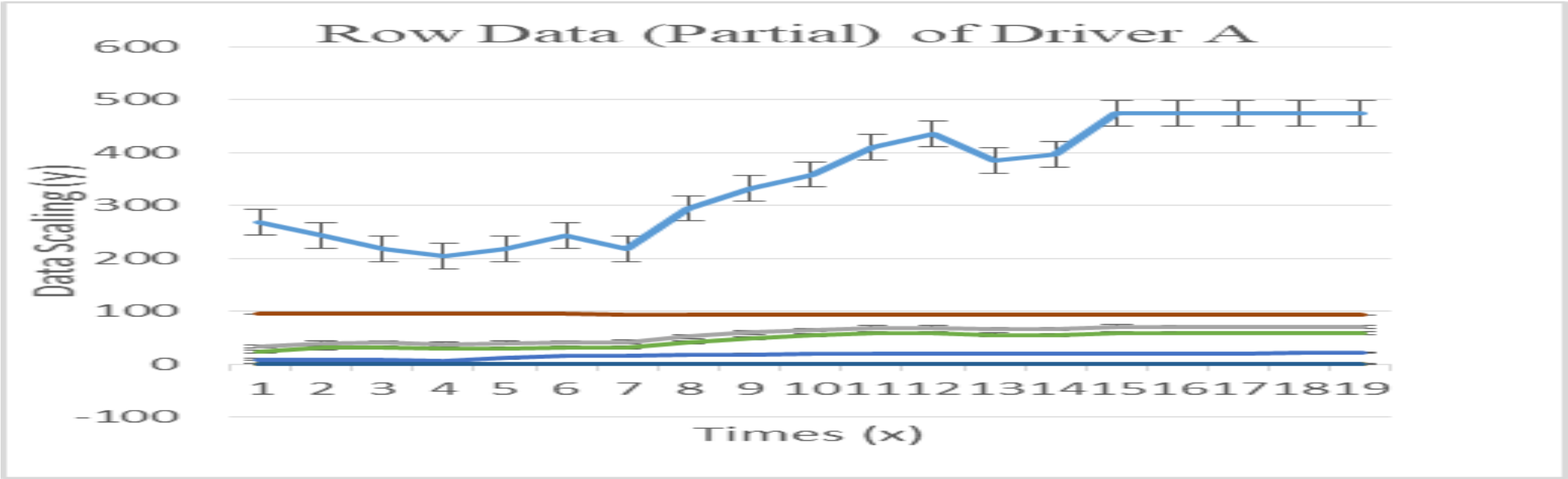

Figure 4 Original data

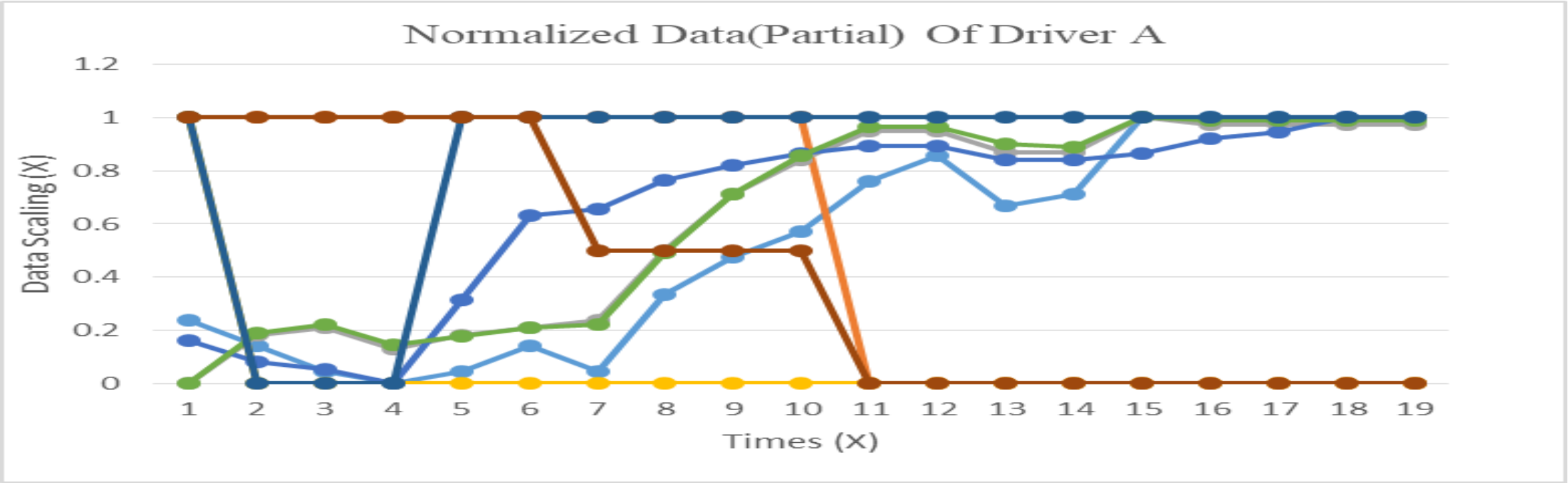

Figure 5 Normalized data

### 4.3 Sliding Window

For distributing data based on time series, the main window and subwindow's length are considered during the time of experiment with the dataset. Every time step, there is a shifting of fixed length subwindow with overlapping, let max time $t_x$ and clearing time of a subwindow is $t_0$, so $t_{0+1}$, $t_{0+2}$ will continue until $t_x$. The windowing process W is expressed as:

$$W = t_0, t_{0+1}, t_{0+2} \ldots . t_x. \tag{5}$$

To find mean, median and standard deviation we drive statistical analysis of our 15 selected features among 51 original features of the dataset and the obtained 15 statistical features. Table 5 shows the moving window, whereas we see a subwindow with orange color, we calculate mean, median and standard deviation from here.

## 4.3 Description of the classifiers

We have considered supervised machine learning classifiers for performing the metrics named *k*NN, SVM, Logistic Regression, REP Tree. The *k*NN (*k*-Nearest Neighbor) is an instance-based traditional machine learning algorithm. Both classification and regression cases *k*-NN can be used and select the number of neighbor through distance calculation of the query points. The equation 6 is used to calculate the Euclidian distance between two points.

$$D = \sqrt{\sum_{i=1}^{n}(X_i - Y_i)^2} \tag{6}$$

SVM stands for Support Vector Machine, used for classification and regression problems. The goal is to find a hyperplane in an N dimensional space and separately classifies the query data point. There is a decision boundary called hyperplane that used to differentiate the classes. It also crates margin separator with the most nearest observations and it performs better if maximize the margin. The equation represents the loss function that indicate maximize the margin.

$$C\,(x, y\, f(x) = \begin{cases} 0, & if\ y * f(x) \geq 1 \\ 1 - y * f(x), & else \end{cases} \tag{7}$$

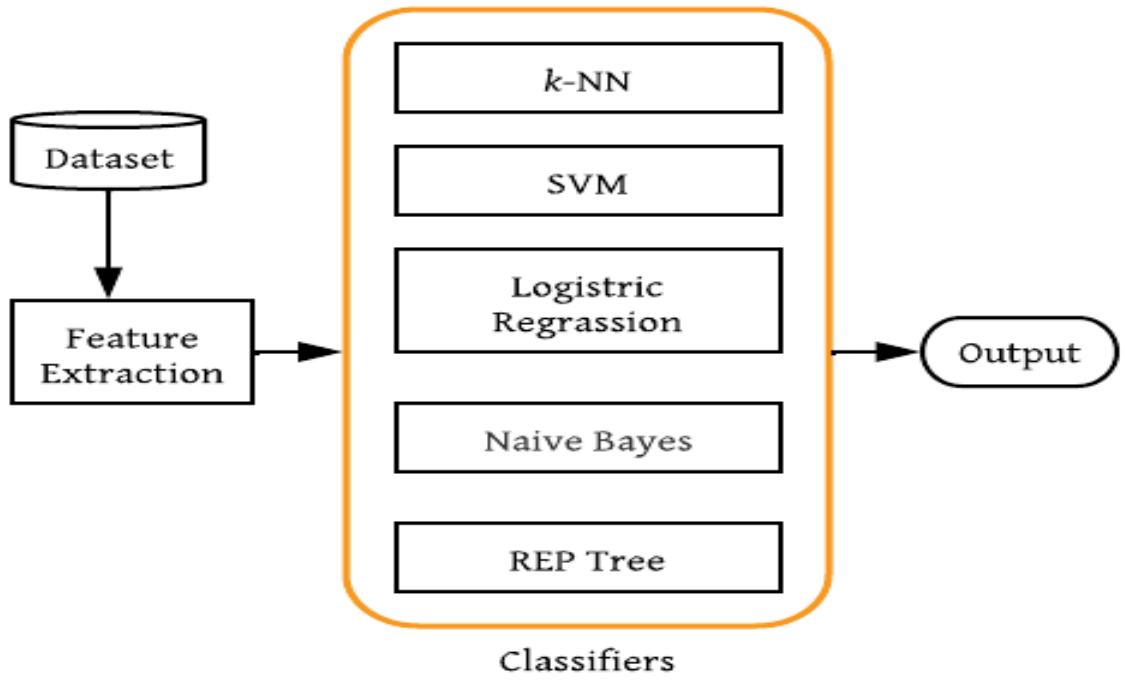

Figure 6 Machine learning approach for this proposed system

Logistic Regression predicts whether something is true of false. Instead of fitting a line to the data, it fits an "S" shaped "logistic function" and the curve goes from 0 to 1.The following equation used to calculate the function, also called sigmoid function.

$$S(x) = \frac{1}{1+e^{-x}} \tag{8}$$

Naïve Bayes is a classifier based on Bayes' theorem. It assumes that the presence of a particular feature in a class is unrelated to the presence of any other feature. Naïve Bayes model is easy to build a large dataset and outperform with sophisticated classification method. The way Naïve Bayes is used to calculate the posterior probability, shows in the equation below.

$$P(c|x) = \frac{P(x|c)P(c)}{P(x)} \tag{9}$$

REP (Reduced Error Pruning) Tree is a classification technique, from a given dataset it generates decision tree. It is seemed to be the extension of the C4.5 by improving the pruning phase. A distinct pruning dataset is used by the method and create multiple trees in different iteration. Finally select the best one. As measure, mean squared error is used for prediction the model by the tree [34]. To find the mean squared error the following equation is used.

$$MSE = \frac{1}{n}\sum_{i=1}^{n}(Y_i - \widehat{Y_i})^2 \quad (10)$$

## 4.4 Performance Metrics

The dataset represented as $X\epsilon R^{Nx*Mx}$ and we selected 15 features from the original dataset of 51 features. The new dataset are express as follows: $X_i$ = X - Σ ($A_h$+$B_i$+$C_s$+$D_c$). We have considered supervised machine learning classifiers to identify the drive behavior. Previous researcher has done some work with the classifiers of e.g. DT (Decision Tree), *k*NN (*k*-Nearest Neighbor), RF (Random Forest), MLP (Multilayer Perceptron) [9] and SVM (Support Vector Machine), RF, *k*NN [23].

In OBD-II the features which are publicly available and also in the Ocslab dataset, we used for preparing confusing metrics. Most of the researchers find accuracy to identify the driver behavior and a few number researchers use precision and f-score [2].

To measure the performance we have used four indicator named accuracy, precision, F-Measure, Recall. The computation and the evaluating performance of the classifiers are occurred through confusion metric. Once the model is generated then the classifier is tested by using test dataset to check the model accuracy. Precision indicate how close or dispersed the measurement is to each other. It measures the number of correct positive predictor made. Recall is a metric that quantifies the number of correct positive predictions made out of all positive predictions that could have been made.

The number of *FP*'s, *FN*'s, *TP*'s, and *TN*'s cannot be calculated directly from this matrix. The values of *FP*'s, *FN*'s, *TP*'s, and *TN*'s for class ***i*** ($1 \le i \le n$) are determined as per [35].

$$TP_i = a_{ii}. \quad (11)$$

$$FP_i = \sum_{\substack{j=1,\\ j\neq i}}^{n} a_{ji}. \quad (12)$$

$$FN_i = \sum_{\substack{j=1,\\ j\neq i}}^{n} a_{ij}. \quad (13)$$

$$TN_i = \sum_{\substack{j=1,\\ j\neq i}}^{n}\sum_{\substack{k=1,\\ k\neq i}}^{n} a_{jk} \quad (14)$$

The final confusion matrix, which has dimension 2 × 2, comprises the average values of the n confusion matrices for all classes. For a binary, i.e. two-class problem, a confusion matrix gives the number of false positives (*FP*'s), false negatives (*FN*'s), true positives (*TP*'s), and true negatives (*TN*'s). From this confusion matrix, accuracy, precision, recall, and $F_1$-score are calculated in the following way:

$$Precision = \frac{TP}{TP + FP} \times 100\% \quad (15)$$

$$Recall = \frac{TP}{TP + FN} \times 100\% \quad (16)$$

$$F_1\text{- score} = \frac{2 \times \ \ precision \times recall}{precision + recall} \times 100\% \tag{17}$$

$$Accuracy = \frac{TP + TN}{(TP + FN) + (FP + TN)} \times 100\% \tag{18}$$

Table 6 Confusion metrics of several classifiers of driver A and D

| Classifier | Accuracy of the model | Accuracy by class (Binary) | | | |
|---|---|---|---|---|---|
| | | Precision | F1-Score | Recall | Class (Driver) |
| Naive Bayes | 96.15 | 92.1 | 95.3 | 98.8 | A |
| | | 97.8 | 98.90 | 97.9 | D |
| Logistic Regression | 98.12 | 97.4 | 97.5 | 98.1 | A |
| | | 98.8 | 98.8 | 99.0 | D |
| *k*-NN | 99.99 | 1.00 | 1.00 | 1.00 | A |
| | | 99.00 | 1.00 | 1.00 | D |
| REP Tree | 99.95 | 1.00 | 99.90 | 99.90 | A |
| | | 1.00 | 1.00 | 1.00 | D |
| SVM | 99.88 | 98.99 | 98.87 | 98.99 | A |
| | | 99.0 | 99.0 | 99.1 | D |
| ZeroR (Baseline) | 78.54 | - | - | 0.0 | A |
| | | 78.0 | 88.8 | 1.0 | D |
| AdaBoost (Ensemble) | 99.91 | 1.00 | 99.9 | 99.8 | A |
| | | 1.00 | 1.00 | 1.00 | D |

Table 7 Confusion metrics of several classifiers of all drivers

| Classifier | Accuracy of the model of full dataset | Accuracy by class (multi class) | | | |
|---|---|---|---|---|---|
| | | Precision | F1-Score | Recall | Class(Driver) |
| Naive Bayes | 29.00% | 41.8% | 37.4% | 33.8% | A |
| | | 33.4% | 20.9% | 15.2% | D |
| KNN | 76.35% | 95.1% | 91.6% | 88.3% | A |
| | | 76.1% | 76.1% | 76.1% | D |
| SVM | 64.00% | 95.5% | 96.1% | 97.3% | A |
| | | 50.6% | 54.3% | 59.6% | D |
| REP Tree | 97.14% | 99.5% | 99.4% | 99.3% | A |
| | | 96.6% | 96.7% | 96.6% | D |
| *ZeroR* (Baseline) | 14.03% | - | - | 0.00 | A |
| | | 14.0% | 24.6% | 1.00% | D |
| AdaBoost (Ensemble) | 20.90% | 70.9% | 79.1% | 89.5% | A |
| | | 15.5% | 26.9% | 1.00% | D |

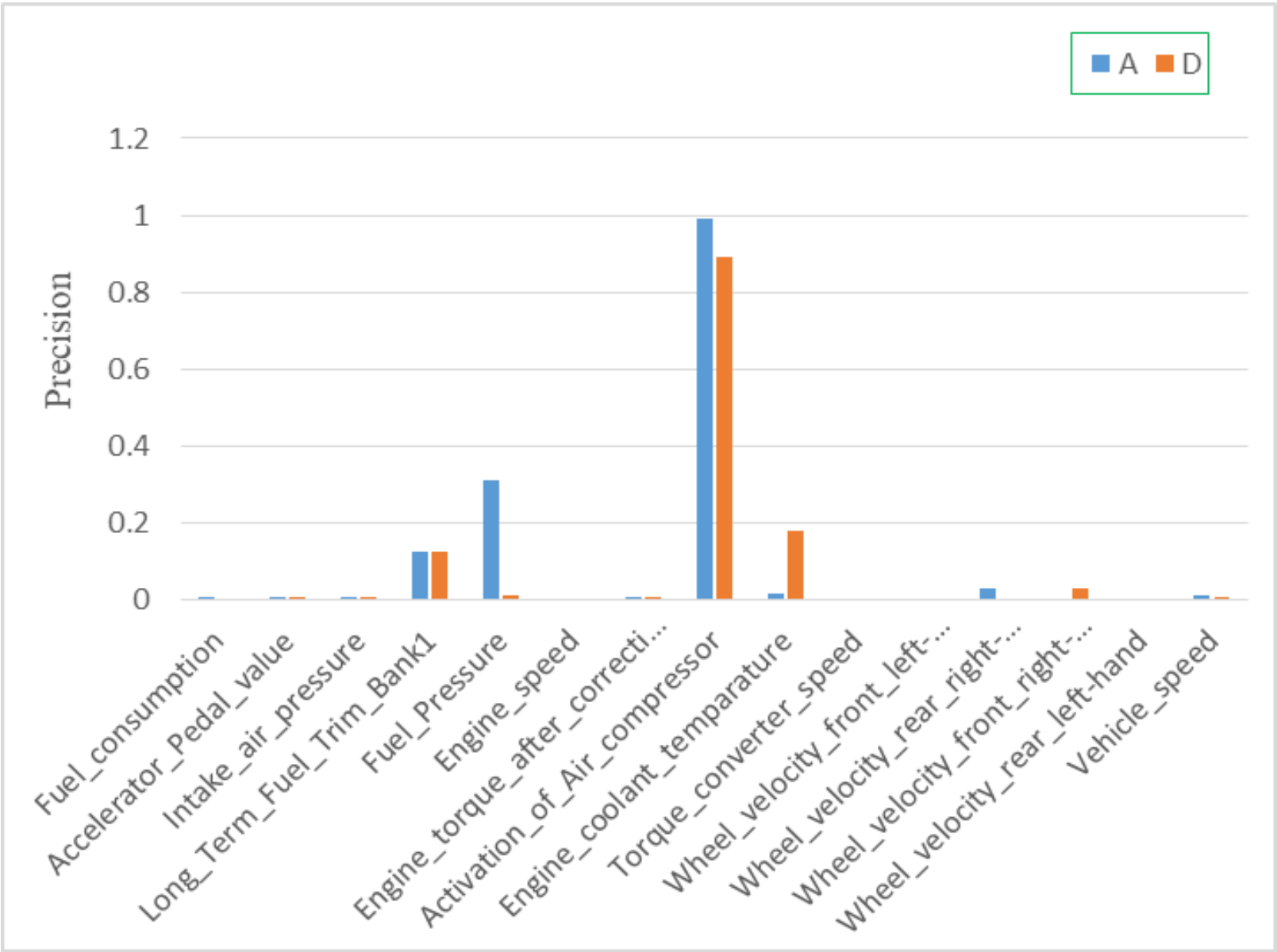

Figure 7. Precision using Naive Bayes model of driver A and D

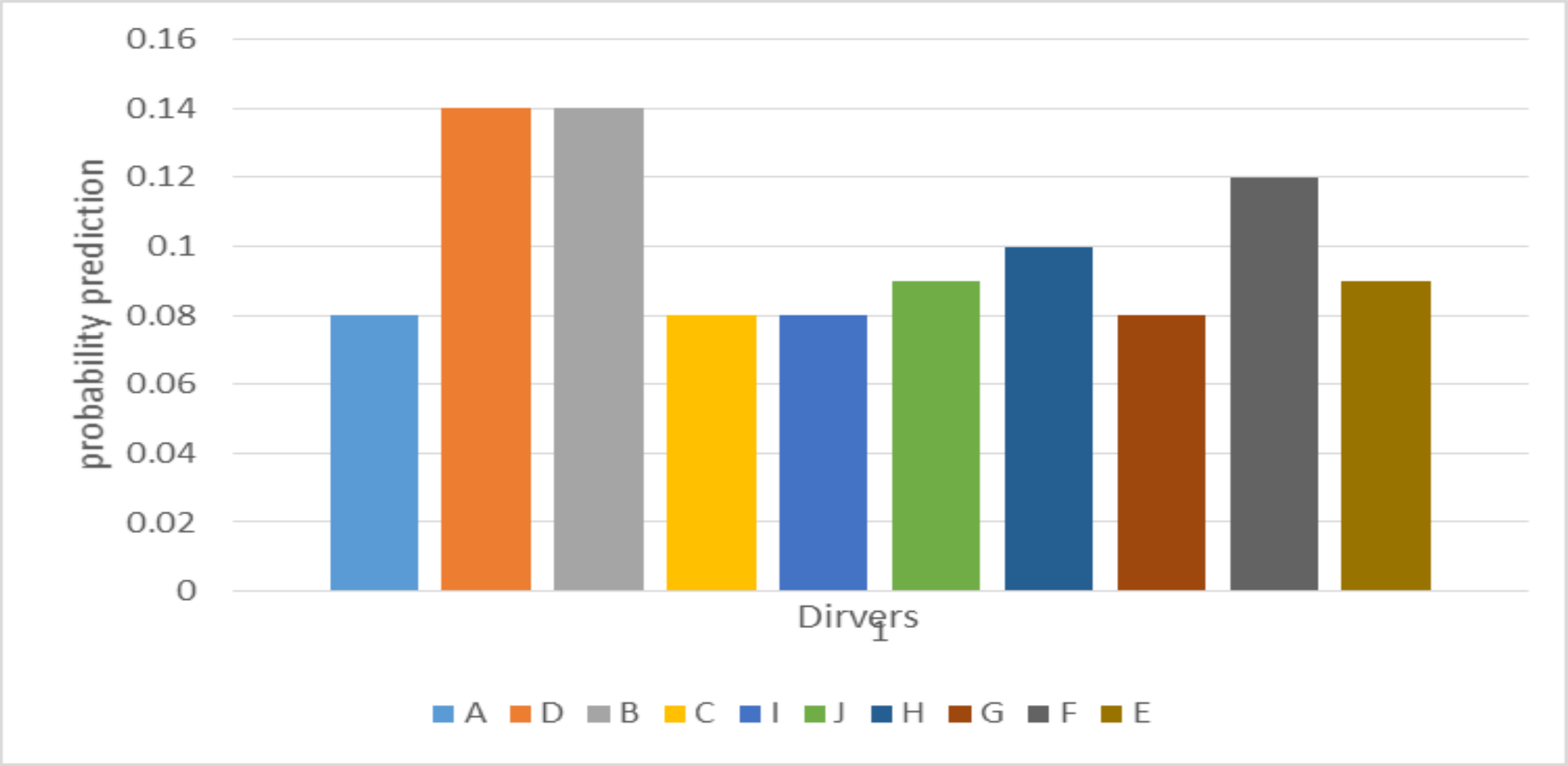

Figure 8. Randomly multi class probability prediction using Naive Bayes

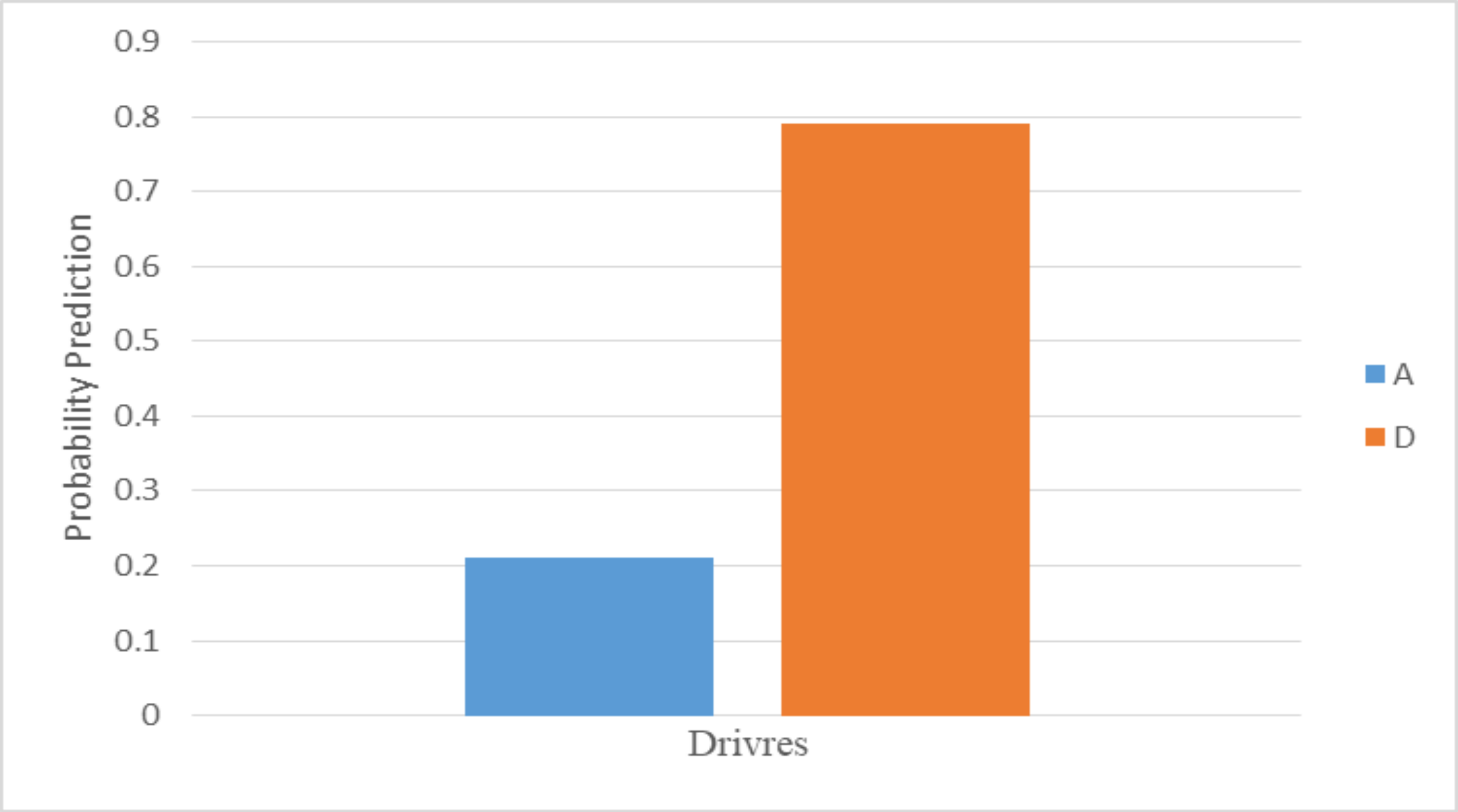

Figure 9. Binary class probability prediction using Naive Bayes

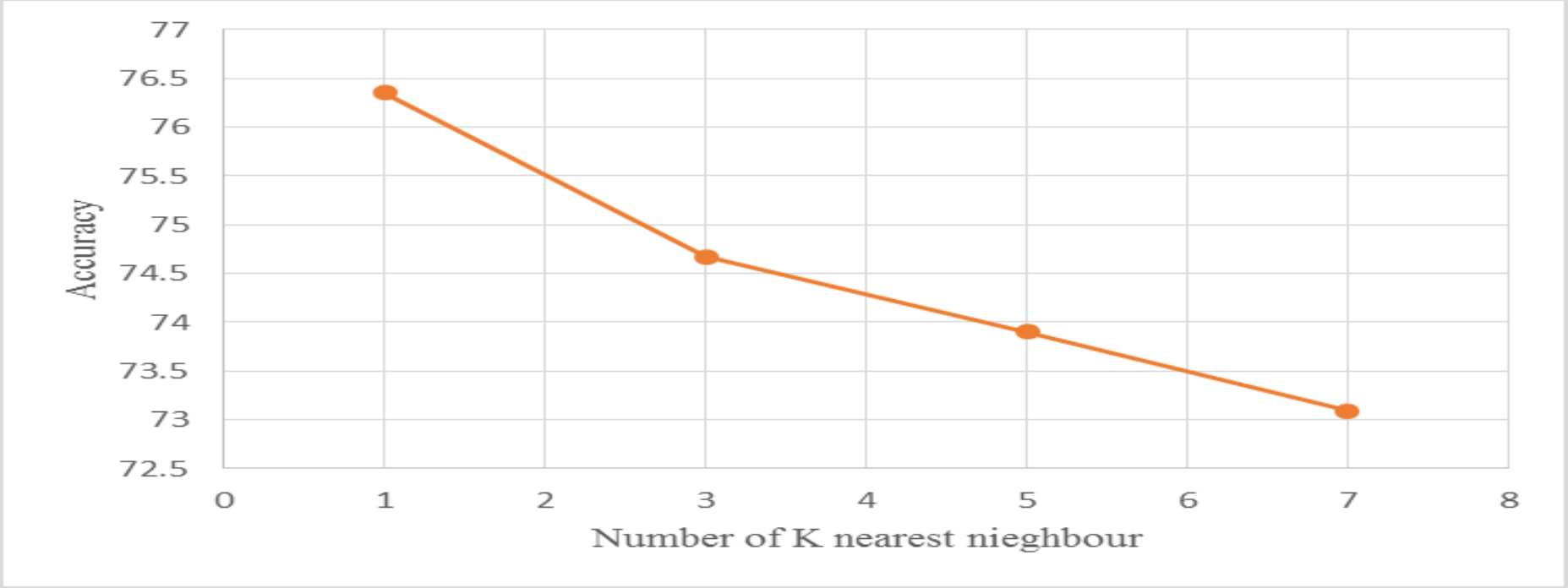

Figure 10 *k*-NN algorithm

## 5. Results and Discussion

To evaluate the generalization of the model we have considered K fold cross-validation for low bias and a modest variance [33]. We have used 10 folds where in each fold contains 9 blocks are used for training and the remaining group taken as a test data set and obtained the mean performance.

For classification of the driver we have introduced the prominent supervised algorithm named Naive Bayes, Logistic Regression, *k*NN, REP Tree and SVM. Table 6 shows the result of all classifiers through confusion metric. Among them *k*NN shows 99.99 % (highest) and Naive Bayes performs 96.15% (lowest) accuracy respectively. Mentionable that we have used 15 features and two drivers among 51 features and 10 drivers respectively from the Ocslab dataset. The results of ensemble classifiers where adaboost and voting are given 99.91% and 60.22% accuracy respectively.

Again we have calculated all the drivers' accuracy using the full dataset. Table 7 is representing the accuracy of only two drivers (A, D) and it also represents the baseline accuracy of 14.03%. In this research we have figured out the *ZeroR* algorithm to calculate the baseline. Moreover, Adaboost uses an ensemble algorithm and the accuracy is important in this research because the other classifier's accuracy is better than this. The model is statistically significant because the accuracy of *k*-NN is better than baseline accuracy.

From the above discussion of Table 6 and 7 we have recognized that state-of-the-art algorithms provide the best accuracy when the driver is less for the Ocslab dataset for public OBD-II in service 01.

Figure 7 illustrates the precision of an algorithm named Naive Bayes, feature *"Accivation_of_Air_Comprssior"* points out the height result for driver A and D. If we calculate all drivers randomly through ZeroR then the baseline accuracy is 14.03 % whereas the Naive Bayes shows 29.93% accuracy on the full Ocslab dataset. This comparison indicates the statistical significance. In the Figure 8 and 9, there is another statistical importance that shows the multi class and binary class probability prediction respectively. If we precisely classify the driver then we must have to consider less number of drivers like Figure 6 shows the good results of driver D than the driver D in Figure 8. The accuracy increases of drive D from 0.14 % to 0.8%, that is more statistically significant.

Tuning the result through the hyper parameter of *k*NN -1, 3, 5, 7 shown in Figure 10 with batch size 100 and Euclidean distance is used for finding the distance function. Each and every hyper parameter gives different results, the higher the number of the nearest neighbor, the lower the accuracy. In REP Tree uses the size of the tree is 1397 with depth and learning rate are 26 and 0.001 accordingly, obtained 99.95 % accuracy for driver A and D.

## 5.1 Comparative Performance Analysis

Table 8 shows the comparative analysis among this work and the works already have done before. The height accuracy 99.99 % belongs to this work based on classifier, application domain and the dataset.

**Table 8 Comparative analysis**

| Method/Work Done | Work Domain | Dataset | No. of Class | Application Domain | Classifier | Accuracy |
|---|---|---|---|---|---|---|
| This work | Driver Identification | Ocslab | 2 (A & D) | Machine Learning | *k*-NN | 99.99% |
| Kwak BI et al. [9] | Driver Profiling | Ocslab | 2 (A & E) | Machine Learning | *k*-NN | 95.70% |
| Wakita T et. al [18],[25], [30],[32] | Driver Identification | Driver signal data | 276 | - | GMM | 76% |
| Enev et al.[23] | Automobile Driver Fingerprinting | In-Vehicle Data | 1 (15 Drivers) | Machine Learning | *k*-NN | 100% |
| Zhang J et al. [35] | Driver Behavior Identification | Ocslab | 2 (B & C) | Deep Learning | LSTM-15 | 99.82% |
| Girma A et. al [27] | Driver identification | Vehicular data trace-2 | 2 (4 Drivers) | Deep Learning | LSTM | 99.00% |
| Choi S et. al[24] | Classification of Driver Behavior | Vehicle signal | 6 | Statistical | Hidden Markov Model (HMM) | 25.00% |
| Ullah S et. al [16] | Lightweight driver behavior identification | Ocslab | - | Deep Learning | GRU | 98.72% |
| Nishiwaki Y et. al [31] | Driver Identificatio | Gas and Brake pedal reading | 276 | - | GMM | 76.00% |
| Xun Y et. Al [26] | Driver Fingerprinting | - | 10 | Deep Leaning | CNN | 100% |

## 7. Conclusions

Driver identification is our prime aim by using telemetric data in terms of best accuracy of the classifiers. The CAN-BUS data was collected through OBD-II. Only public PIDs are used for this research work because some non-public PIDs (which are hard to identify) are available in OBD-II. Previous researchers also use the same PIDs for the several classifiers. Logistic Regression is a prominent supervised learning. We could not build the model by using the full Ocslab dataset, even it was a 24 hours continuous process. Partial dataset with two drivers we have built the model successfully and achieve good accuracy. We have achieved statistical significance because the baseline classifier is smaller than the others. Moreover, the accuracy of the ensemble and other supervised learning classifiers are almost the same. In *k*NN classifiers, there is a computational complexity e.g. it takes more processing time when we consider the higher nearest neighbor to build the model and provide less accuracy, in contrast, we have found height accuracy with lower nearest neighbor and less computational complexity. To identify the driver we need around 100 % accuracy. In this regard, *k*NN shows the height accuracy of 99.99% among two drivers with 15 features. Whereas for the full data set with 10 drivers the accuracy is 76.36 % which is unsuitable for driver identification.

**List of Abbreviations**

CAN- Controller Area Network
DTC- Diagnostic Trouble Code
ECU- Electronic Control Unit
OBD- On-Board Diagnostic

**Declaration**

Availability of data and material: Ocslab driving dataset, https://ocslab.hksecurity.net/Datasets/driving-dataset
Competing interests: Find out outperform accuracy of the model.
Funding: Not applicable
Authors' contributions: Found best model compare with different classifiers
Acknowledgements: We thank KIA Motors Corporation in South Korea for provided us the dataset.